\documentclass{article}

\PassOptionsToPackage{numbers,compress}{natbib}
 \usepackage[preprint]{neurips_2026}

\usepackage[utf8]{inputenc} 
\usepackage[T1]{fontenc}    
\usepackage{hyperref}       
\usepackage{url}            
\usepackage{booktabs}       
\usepackage{amsfonts}       
\usepackage{nicefrac}       
\usepackage{microtype}      
\usepackage{xcolor}         
\usepackage{comment}
\usepackage{bbm}
\usepackage{array}

\usepackage{lipsum}  
\usepackage{xspace}
\usepackage{amsmath}
\usepackage{accents}
\usepackage{cleveref}
\usepackage{enumitem}
\usepackage{graphicx}
\usepackage{multirow}
\usepackage{graphicx}
\usepackage{booktabs}
\usepackage{makecell}
\usepackage{tabularx}
\usepackage{algorithm}
\usepackage{algorithmic}

\graphicspath{{Figures/}}

\newlength{\dhatheight}
\newcommand{\doublehat}[1]{%
    \settoheight{\dhatheight}{\ensuremath{\hat{#1}}}%
    \addtolength{\dhatheight}{-0.35ex}%
    \hat{\vphantom{\rule{1pt}{\dhatheight}}%
    \smash{\hat{#1}}}}
\newcommand{\method}{\textsc{MoMo}\xspace}

\title{MoMo: Conditioned Contrastive Representation Learning for Preference-Modulated Planning 
}

\author{%
  Yusuf Syed, Viraj Parimi, Brian Williams \\
  Massachusetts Institute of Technology \\
  Cambridge, MA 02139 \\
  \texttt{\{yusufs,vparimi,williams\}@mit.edu}
}

\begin{document}

\maketitle

\setlength{\textfloatsep}{16pt plus 2pt minus 0pt} 

\begin{abstract}
Temporally contrastive representation learning induces a latent structure capable of reducing long-horizon planning to inference in a low-dimensional linear system. However, existing contrastive planning work learns a single latent geometry which cannot distinguish multiple valid behaviors trading task efficiency against risk exposure for the same start-goal query. We introduce \method, a \emph{preference-conditioned} contrastive planner allowing a scalar user preference to continuously modulate plan conservativeness at inference time, without retraining. \method learns a joint conditioning of the representation geometry and latent prediction operator via Feature-Wise Linear Modulation and low-rank neural modulation, respectively. We show that our formulation preserves the probability density ratio encoded in the representation space that is required for inference-driven contrastive planning, further retaining its inference-time efficiency. 
Across six environments, \method smoothly adapts plan safety according to user preferences, yielding improved temporal and preferential consistency over state augmentation baselines.
%

\end{abstract}

\section{Introduction}
\label{sec:intro}

\begin{figure*}[t]
    \centering
    \includegraphics[width=\textwidth]{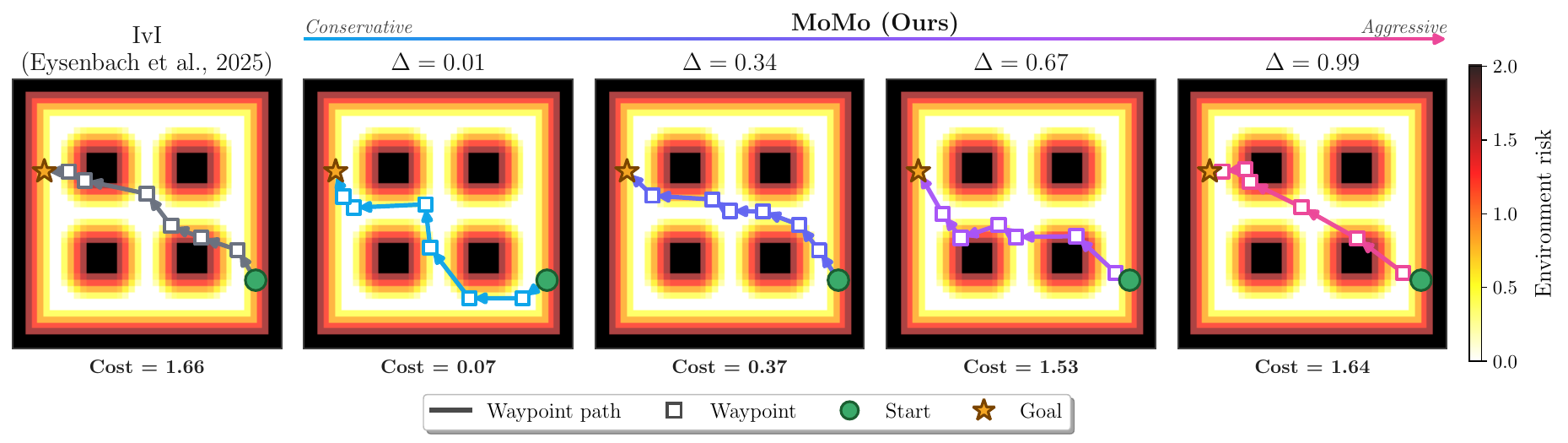}
    \vspace{-0.25cm}
    \caption{Preference-conditioned planning in a risk-structured environment with four obstacles and a cost function reflecting obstacle proximity. Different scalar safety preferences ($\Delta$) modulate waypoint plans, 
    ranging from low-risk detours to more direct, riskier routes. 
    }
    \label{fig:teaser}
\end{figure*}

Robotic systems are increasingly deployed in safety-critical domains such as healthcare and disaster response, where behaving either too aggressively or too conservatively can be costly~\citep{ebar, disaster_response_robotics, STAR}. For instance, a surgical robot may maintain a larger clearance during a delicate phase of an operation, but favor a more direct maneuver when rapid intervention is required. In these settings, the same start and goal can admit multiple viable trajectories that trade task efficiency against exposure to risk, as illustrated in Figure~\ref{fig:teaser}. This balance is not fixed \textit{a priori}, but varies with mission context, ultimately reflecting a specified safety-efficiency tradeoff for planning behavior. This creates a need for \emph{preference-conditioned} planning, in which a planner flexibly reshapes its behavior at inference time, without retraining, in response to a scalar safety preference encoding the user's risk tolerance.

Existing approaches to risk-aware planning span model-based methods that depend on known dynamics~\citep{ccpc_blackmore_ono, ono-ira, scora, groot, online_rbmp_dynamic_environments, ccmp_high_dimensional}, learning-based methods with fixed risk specifications~\citep{viraj_gcrl, rail}, and recent risk-conditioned policies that allow inference-time adaptation~\citep{rcnmp, Yoo_Park_Woo_2024, saformer, pacer}. Despite this progress, the intersection of inference-time preference control, planning under unknown dynamics, and scalability to high-dimensional or long-horizon settings remains underexplored.

Contrastive Representation Learning for Planning (CRLP) offers a promising foundation for addressing this gap~\citep{contrastive_rl, ivi}. By learning representations from trajectory data whose geometry reflects temporal reachability, CRLP induces a latent structure reducing long-horizon planning to low-dimensional linear inference~\citep{ivi}, and has shown practical viability in robotic settings~\citep{zheng2025contrastivedifferencepredictivecoding, zheng2025stabilizingcontrastiverltechniques}. This makes CRLP attractive when dynamics are unknown and efficient inference-time planning is required. Existing contrastive planning methods, however, are preference-agnostic, encoding temporal similarity alone and recovering only a single mode of behavior for a given task. 
Extending CRLP to support preference-conditioning is nontrivial as a preference-conditioned variant must preserve this structure while allowing it to vary meaningfully with a user-specified preference. We approach this challenge by considering behavioral differences arising at two levels: the organization of states in the representation space, and the transitions between them. Conditioning representations yields a preference-aware geometry, while conditioning the transition operator reshapes how trajectories evolve under varying preferences. 

To unify these components, we introduce \method, a preference-conditioned contrastive planning framework that jointly conditions the representation geometry and the transition operator, enabling inference-time behavioral modulation 
while preserving the planning structure of CRLP. Training \method requires trajectory data with diverse behavioral coverage. 
Therefore, we curate training datasets offering richer coverage of the joint state-preference space, for this purpose. 
In summary, our contributions are:

\begin{enumerate}[itemsep=1.5pt, leftmargin=*, topsep=1.5pt]
    \item \textbf{Problem formulation.} We contextualize preference-conditioned planning within the CRLP framework and identify a conditioning scheme that preserves its inference-driven structure.
    \item \textbf{Architecture.} We introduce \method, which modulates planning behaviors by jointly conditioning the encoder and the latent transition operator.
    \item \textbf{Empirical results.} We show that \method enables behavioral modulation of plans in response to user preference across six environments spanning a diverse set of tasks, while retaining the inference-time planning efficiency.
    
\end{enumerate}

\section{Related Work}
\label{sec:related-work}

\paragraph{Model-Based Risk-Aware Planning.}
Classical model-based methods provide explicit safety mechanisms but rely on known dynamics. Chance-constrained trajectory optimization enforces user-specified bounds on failure probability, allocating risk across the planning horizon~\citep{ccpc_blackmore_ono, ono-ira, scora, groot, online_rbmp_dynamic_environments, ccmp_high_dimensional}. Risk-aware control bounds tail risk through conditional value-at-risk (CVaR) constraints~\citep{cvar_opti}, while control barrier functions filter control inputs to maintain safety invariants~\citep{cbf}. These approaches enable principled safety guarantees but assume known, often linearized dynamics and become expensive in high-dimensional or long-horizon settings, motivating learning-based alternatives that remove the dependence on explicit dynamics models. 

\paragraph{Adaptive Safe Policy Learning.} Learning-based methods relax dependence on known dynamics by extracting behaviors directly from trajectory data. \citet{viraj_gcrl} combine graph-based waypoint planning with safe goal-conditioned RL policies for multi-agent navigation, but enforce safety through graph pruning, which can disconnect goals beyond high-risk regions~\citep{parimi2026riskboundedmultiagentvisualnavigation}. \citet{rcnmp} and \citet{Yoo_Park_Woo_2024} introduce risk-conditioned policies that allow inference-time adaptation across a continuous risk spectrum, but rely on direct policy conditioning, with limited scalability over longer horizons. 
In the offline setting, decision-transformer approaches such as \citet{cdt} condition on desired return and constraint thresholds for zero-shot adaptation across cost budgets, with extensions to safety verification~\citep{saformer}, signal temporal logic specifications~\citep{sdt}, and constraint-conditioned actor-critic learning~\citep{ccac}. Benchmark suites such as DSRL~\citep{fsrl} have accelerated progress in this area, but center on safety-constrained control rather than granular preference modulation. \citet{pacer} follow a different approach, learning preference-conditioned terrain costmaps for downstream planning. Across these methods, preference or risk adaptability operates at the level of policies or costmaps, not through a learned representation whose structure supports planning by inference.

\paragraph{Preference-Conditioned and Multi-Objective RL.}
Multi-objective RL (MORL) addresses tradeoffs by learning policies or value functions conditioned on preferences over vector-valued returns~\citep{yang2019morl, dpmorl}. 
Offline variants such as PEDA, together with the D4MORL benchmark, extend this to preference-conditioned decision transformers trained on annotated demonstrations~\citep{peda}, while recent work emphasizes comprehensive coverage of the Pareto front for unseen preference adaptation~\citep{cmorl}. These methods cast the problem as optimization over vector-valued returns, whereas our setting studies how different safety preferences reshape plans for a fixed start-goal pair.

\paragraph{Latent Planning.}
Beyond policy-level adaptation, learned representations offer a route to tractable planning in high-dimensional settings. Early work learns latent dynamics from observations for control and online planning~\citep{watter_e2c, hafner_planet}, with subsequent methods studying planning directly in latent spaces~\citep{ichter2018robotmotionplanninglearned, park2024latentplanning, Zhang_2025}. \citet{laplass} extend this to stochastic, risk-aware planning by learning linear dynamics in latent space and propagating uncertainty through probabilistic flow tubes~\citep{dong2012pft} sampled from a variational autoencoder~\citep{Kingma_2019}. These approaches highlight the value of learned representations for planning under unknown dynamics yet do not condition them on user preferences.

\paragraph{Contrastive Planning and Conditional Representation Learning.}
Our work builds on contrastive representations for planning. Time-contrastive networks and successor representations demonstrate how temporal structure in trajectory data can be captured in learned representations~\citep{tcn, successor_representation, successor_features}. \citet{contrastive_rl} show that contrastive learning can be interpreted as goal-conditioned RL, with subsequent work improving performance from offline data~\citep{zheng2025contrastivedifferencepredictivecoding, zheng2025stabilizingcontrastiverltechniques}. \citet{ivi} show that temporally contrastive representations induce a Gauss-Markov latent structure in which long-horizon planning reduces to efficient inference over intermediate waypoints in a lower-dimensional linear system. Separately, conditional representation learning offers tools for reshaping embeddings as a function of auxiliary variables. Examples include masked subspaces~\citep{csn}, kernel-based objective reweighting~\citep{fair_ccl, tsai2022conditionalcontrastivelearningkernel, ccn}, Feature-wise Linear Modulation (FiLM)~\citep{film} for controlling intermediate activations, and HyperNetworks~\citep{hypernetworks} for generating network parameters from auxiliary inputs. These conditioning methods must be applied carefully so as not to disrupt the latent structure enabling inference-based planning, as analyzed in Section~\ref{sec:conditioning-prelim}.
\section{Problem Setup and Preliminaries}
\label{sec:preliminaries}

\subsection{Preference-Conditioned Planning from Trajectory Data}
\label{sec:problem-setup}

Consider a system with state space $\mathcal{S} \subseteq \mathbb{R}^d$ and per-step cost $\delta_t \in \mathbb{R}_{\geq 0}$ penalizing proximity to constraints such as obstacles. Preference-conditioned planning seeks a learned mapping $\mathcal{P}$ that, given a query $\mathbf{q} = (s_0, s_g, \Delta)$ with $s_0, s_g \in \mathcal{S}$ and preference $\Delta \in [0, 1]$, returns a sequence of $n$ waypoints $s_{1:n} = \mathcal{P}(\mathbf{q})$ from $s_0$ to $s_g$.
$\Delta$ specifies the desired safety behavior, which we quantify using a CVaR-based measure in Section~\ref{sec:preferences}. The planner is trained from an offline dataset
\begin{equation}
    \mathcal{D} = \bigl\{ (\tau_i, \Delta_i) \bigr\}_{i=1}^N,
    \quad 
    \tau_i = \bigl(s_t^i, \delta_t^i\bigr)_{t=0}^{T_i-1},
\label{eq:dataset}
\end{equation}
where each trajectory is collected under an unknown behavioral policy and annotated with the scalar preference label $\Delta_i \in [0, 1]$ summarizing the trajectory's safety. 
We require the preference-conditioned planner $\mathcal{P}$ to satisfy three properties:
\begin{enumerate}[itemsep=1.5pt, leftmargin=*, topsep=1.5pt]
    \item \textbf{Temporal consistency.} Waypoints correspond to dynamically feasible short-horizon transitions.
    \item \textbf{Preference alignment.} For a fixed $(s_0, s_g)$, varying $\Delta$ should modulate behavior such that lower values of $\Delta$ yield plans whose induced trajectory exhibits lower risk.
    \item \textbf{Efficient inference.} $\mathcal{P}$ produces preference-conditioned plans $s_{1:n}$ at test time without retraining, and the computational cost of producing that plan scales gracefully in $n$.
\end{enumerate}
These requirements motivate a planning-oriented latent representation learned directly from trajectory data, whose geometry captures temporal reachability and can be queried efficiently at inference time.

\begin{figure*}[t]
    \centering
    \includegraphics[width=\textwidth]{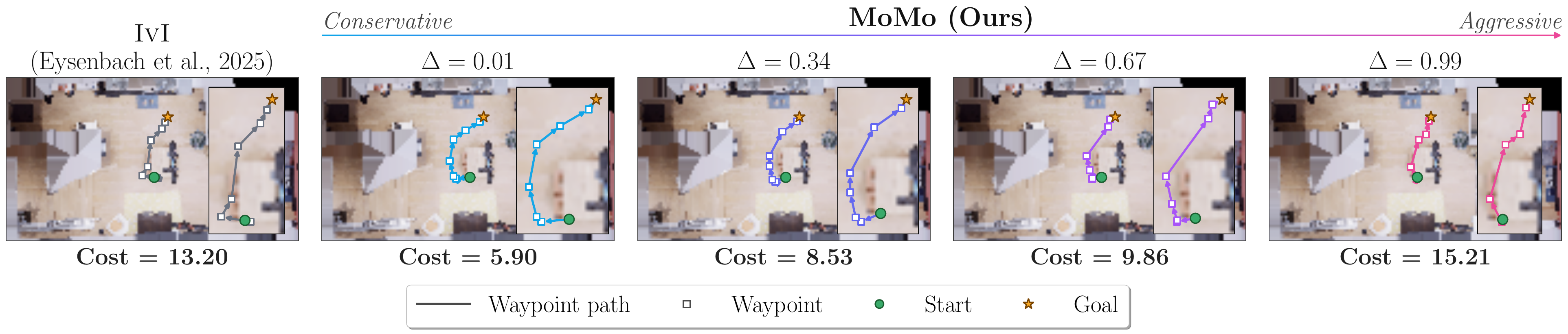}
    \vspace{-0.25cm}
    \caption{Preference-conditioned planning on the 29-dimensional \textit{Ant} agent in an AI Habitat environment. Risk is defined as a linear function of obstacle proximity within an influence radius. For a fixed start and goal, increasing $\Delta$ reduces the planned obstacle clearance for a more efficient path. 
    }
    \label{fig:ant_headline}
\end{figure*}

\subsection{Contrastive Planning from Trajectory Data}
\label{sec:contrastive-prelim}

Our method builds on CRLP~\citep{contrastive_rl, ivi}, which learns a latent space whose geometry reflects temporal reachability. Given an anchor state $s$, a positive future $s_+$ is sampled from the discounted state-occupancy distribution $p_{\gamma}(s_+\vert  s)$~\citep{contrastive_rl}, using a discounting factor $\gamma\in(0,1)$ which controls the effective temporal horizon of the sampling distribution.
Negative pairs are formed from an anchor and negative future $s_-$ by sampling independently from the product of marginal distributions $p(s)p(s_-)$.

Let $\psi\colon \mathcal{S} \to \mathbb{R}^{d_z}$ and $\phi\colon \mathcal{S} \to \mathbb{R}^{d_z}$ denote future and anchor encoders, respectively, with representation dimension $d_z$. For a batch of $B$ positive pairs $\{(s^i, s^i_+)\}_{i=1}^B$, we define the logits 
\begin{equation}
    S_{ij} = \exp \Bigl(-\tfrac{1}{2} \bigl\| \phi(s^i) - \psi(s^j_+) \bigr\|^2_2 \Bigr),
\label{eq:logits}
\end{equation}
with diagonal matrix entries scoring positive pairs and off-diagonal entries scoring bootstrapped negative pairs~\citep{ivi}. 
The symmetrized InfoNCE objective~\citep{infoNCE, ivi} trains the encoders to distinguish the positive from negative pairs in the sampled batch:
\begin{equation}
    \mathcal{L}_{\mathrm{NCE}} = -\sum_{i=1}^B \left[ \log\left( \frac{S_{ii}}{\sum_{j=1, \; j\neq i}^B S_{ij}}\right)+ \log\left( \frac{S_{ii}}{\sum_{j=1, \; j\neq i}^B S_{ji}} \right)\right].
\label{eq:infonce}
\end{equation}

Optimizing~\eqref{eq:infonce} pulls temporally related states together and pushes unrelated states apart, so that latent distance reflects temporal reachability. While standard contrastive learning constrains representations to the unit hypersphere~\citep{contrastive_rl}, following~\citet{ivi} we instead bound the expected norm $\frac{1}{d_z}\,\mathbb{E}_{s \sim p(s)} \bigl[ \| \psi(s) \|^2_2 \bigr] \leq c$, enforced via a regularization term and dual gradient descent, where $c$ is the constraint target~\citep{ivi}. The anchor encoder is parameterized as a linear transformation of the future encoder, such that $\phi(s) = A\psi(s)$, where $A \in \mathbb{R}^{d_z \times d_z}$ acts as a latent predictor mapping representations to their $\gamma$-discounted future. Crucially, the optimal logits learned under~\eqref{eq:infonce} encode the density ratio
\begin{equation}
    S_{ij}^* \propto \frac{p_{\gamma}(s_+^j \mid s^i)}{p(s_+^j)}.
\label{eq:density-ratio}
\end{equation}
The representation space learned in~\citep{ivi} therefore encodes how likely it is to see a certain future state given a starting state.
Combined with the assumption that the marginal distribution of representations is isotropic Gaussian and the linear parameterization of $\phi(s)$ in terms of $\psi(s)$, this property induces a Gauss-Markov latent structure that supports tractable planning. \citet{ivi} derive that the resultant latent dynamics from $z = \psi(s)$ to its $\gamma$-discounted future representation $z'$ become
\begin{equation}
    p(z' \mid z) = \mathcal{N} \Bigl( \tfrac{c}{c+1}Az, \; \tfrac{c}{c+1}\,I \Bigr).
\label{eq:latent-dynamics}
\end{equation}
A planning query between start and goal representations $z_0, z_g$ is solved by inferring intermediate latent waypoints $z_{1:n}$ from the Markov chain-factorized Gaussian posterior $p(z_{1:n} \mid z_0, z_g)$. Under~\eqref{eq:latent-dynamics}, this posterior has mean $\mu = \Lambda^{-1} \eta$ and covariance $\Sigma=\Lambda^{-1}$, where $\Lambda$ is a block-tridiagonal matrix defined by $A$ and $c$, and $\eta$ encodes start and goal embeddings~\citep{ivi}, defined in Appendix~\ref{appendix:contrastive-planning}. Long-horizon planning therefore reduces to solving a sparse $nd_z$-dimensional linear system in $\mathcal{O}(nd_z^3)$ time rather than optimizing trajectories in the original observation space. Following~\citep{ivi}, the inferred latent waypoints $z_{1:n}$ are decoded to the observation space plan $s_{1:n}$ by nearest-neighbor lookup in a held-out set $\mathcal{D}_{\mathrm{held}} \subset \mathcal{D}$, mapping each latent waypoint to the observation $s \in \mathcal{D}_{\mathrm{held}}$ whose representation $\psi(s)$ is closest.

\subsection{Preference Conditioning in Contrastive Planners}
\label{sec:conditioning-prelim}

Extending contrastive planning to support preference conditioning requires care. The Gauss-Markov planning structure depends on the density-ratio property of learned logits in~\eqref{eq:density-ratio}, and any preference-conditioned extension must preserve this property. Consider a conditioned logit mirroring~\eqref{eq:logits} with preference-dependent encoders $\phi_\Delta$ and $\textcolor{blue}{\psi_\Delta}$, 

\begin{equation}
    S^{\Delta}_{ij} = \exp \Bigl( -\tfrac{1}{2} \bigl\| \phi_{\Delta}(s^i) - \textcolor{blue}{\psi_{\Delta}}(s^j_+) \bigr\|^2_2 \Bigr).
\label{eq:conditioned-logit}
\end{equation}

If the anchor encoder is parameterized as $\phi_\Delta(s) = \textcolor{red}{A_\Delta}\textcolor{blue}{\psi_\Delta}(s)$, then conditioning can act through two distinct mechanisms. It can deform the representation geometry via $\textcolor{blue}{\psi_\Delta}$, and it can modify how trajectories evolve through that space via the latent transition matrix $\textcolor{red}{A_\Delta}$. We show in Appendix~\ref{appendix:density-ratio-assumption} that this preference conditioning at the level of the logits preserves the density-ratio property, justifying the joint-conditioning architecture designed for \method.


\section{\method: Preference-Conditioned Contrastive Planning}
\label{sec:methods}

We now introduce \method, a preference-conditioned contrastive planning framework built on CRLP. Given an offline trajectory dataset, \method learns a conditioned encoder and prediction matrix that jointly reshape the planning geometry using
the queried safety preference, while preserving the density-ratio structure required for planning by inference. The remainder of this section defines the trajectory-level preference statistic (Section~\ref{sec:preferences}), the conditioning mechanisms (Sections~\ref{sec:FiLM}, \ref{sec:LRNM}), and the training objective (Section~\ref{sec:objective}). Algorithm~\ref{alg:momo} summarizes the training and inference pipeline.

\subsection{Safety Preference from Trajectory Costs}
\label{sec:preferences}
For preference-conditioned planning, we summarize each trajectory's safety behavior with a scalar statistic. Given a trajectory $\tau_i$, let $(\delta_t^i)_{t=0}^{T_i - 1}$ denote the sequence of per-step costs, where $\delta_t^i \in \mathbb{R}_{\geq 0}$ penalizes constraint exposure such as obstacle proximity. Zero-cost segments arise in both conservative and risk-tolerant behavior whenever the agent follows an efficient path through safe regions, so averaged costs fail as a faithful summary by diluting the preference signal.
We therefore use the empirical cost distribution upper-tail, following prior work in risk-conditioned planning that treats tail-risk metrics as a natural conditioning signal~\citep{rail, rcnmp}. Letting $q_\alpha(\tau_i)$ denote the empirical $\alpha$-quantile of $(\delta_t^i)_{t = 0}^{T_i - 1}$, and $\mathcal{T}_\alpha(\tau_i) = \left\{ t \in \{0, \ldots, T_i - 1\}: \delta_t^i \geq q_\alpha(\tau_i) \right\}$ index the upper-tail steps, we define the raw tail-risk statistic as the empirical CVaR~\citep{Rockafellar2000OptimizationOC} of the per-step costs,
\begin{equation}
    \widetilde{\Delta}_i = \mathrm{CVaR}_\alpha(\tau_i) = \frac{1}{|\mathcal{T}_\alpha(\tau_i)|} \sum_{t \in \mathcal{T}_\alpha(\tau_i)} \delta_t^i,
\label{eq:trajectory-cvar}
\end{equation}
This averages over the highest cost portions of the trajectory, isolating behaviorally meaningful interactions with risk while excluding the zero-cost segments common to both conservative and risk-tolerant behaviors. We wrap these raw values $\{\widetilde{\Delta}_i\}$ to lie in $[0, 1]$ across the training dataset, so that $\Delta_i = 0$ corresponds to the most risk-averse trajectory and $\Delta_i = 1$ to the most risk-tolerant. Each trajectory thus carries a single preference label, shared across all states in the trajectory, which sets the preference context for every contrastive pair sampled from it. We assume that $\mathcal{D}$ comprises trajectories that do not take on additional risk without commensurate efficiency benefit. This makes the cost-computed $\Delta$ a meaningful conditioning variable for behavioral modulation.

\begin{figure*}[t]
    \centering
    \includegraphics[width=\textwidth]{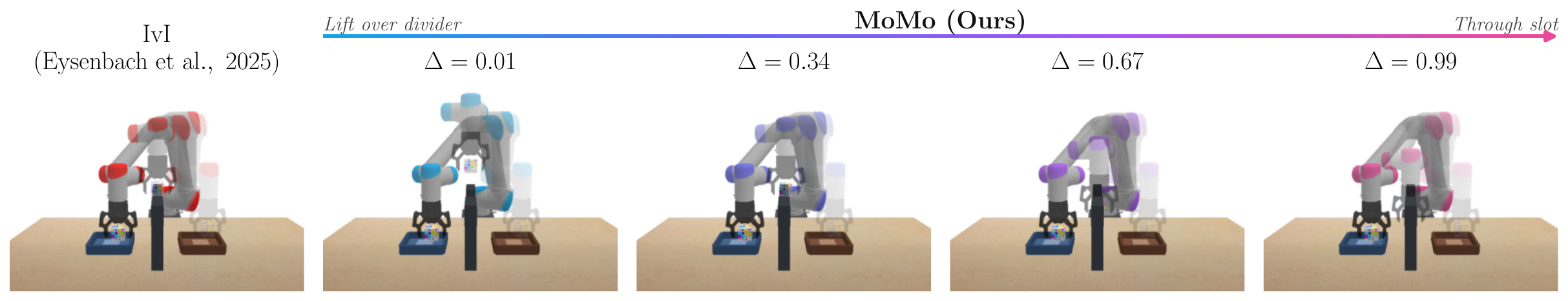}
    \vspace{-0.25cm}
    \caption{Preference-conditioned planning on a pick-and-place \textit{Manipulation} task, where the UR5 robot must move a cube between start and goal regions over a gate divider. \method shows behavioral gradation between safely lifting above and efficiently sliding through the gate as $\Delta$ increases.
    }
    \label{fig:ur5_headline}
\end{figure*}

\subsection{Preference-Conditioned Latent Geometry}
\label{sec:FiLM}

In CRLP, the encoder organizes states according to temporal reachability alone. Under preference-conditioning, the same state may play different roles depending on the desired safety level. A state near an obstacle, for example, may be a plausible waypoint under a risk-tolerant preference but an unlikely one under a conservative preference. The encoder must therefore jointly capture temporal reachability and preferential similarity. Schemes that apply a condition-dependent mask to a shared embedding~\citep{csn} leave the intermediate features unchanged, limiting the expressivity with which preference can reshape the representation. 

We instead modulate intermediate representations using Feature-Wise Linear Modulation (FiLM)~\citep{film}, which applies $\Delta$-dependent affine transformations to the encoder's hidden layer activations, allowing the preference to reshape features hierarchically through the network. Let $h^{(l)}$ denote the pre-activation at layer $l$, before the nonlinearity. A FiLM generator $F_\xi$ with parameters $\xi$ takes input $\Delta$ and produces scale and shift vectors $\{(\lambda_\Delta^{(l)}, \beta_\Delta^{(l)})\}$ for each encoder layer $l$, modulating the activation
\begin{equation}
    a_\Delta^{(l)} = \mathrm{swish}\!\left(\lambda_\Delta^{(l)} \odot h^{(l)} + \beta_\Delta^{(l)} \right),
\label{eq:film-layer}
\end{equation}
where $\odot$ denotes an elementwise product. We use the Swish nonlinearity~\citep{swish} following \citet{ivi}. The scale $\lambda_\Delta^{(l)}$ controls the strength of each feature channel under the queried preference, while the shift $\beta_\Delta^{(l)}$ adjusts each feature's bias, effectively gating its contribution to downstream layers.
Applying this modulation across all encoder layers defines the preference-conditioned future-state encoder $\textcolor{blue}{\psi_\Delta}(s)$ with FiLM parameters generated by $F_\xi(\Delta)$.

\subsection{Preference-Conditioned Latent Predictions}
\label{sec:LRNM}

To support varying latent transition dynamics, we parameterize a preference-dependent prediction matrix $\textcolor{red}{A_\Delta}$ as a shared base matrix with a low-rank preference-dependent perturbation. This reflects the inductive bias that dominant temporal structure is shared across preferences, while conditioning shifts which transitions are locally favored. 
Let $A_0 \in \mathbb{R}^{d_z \times d_z}$ denote the shared matrix, and $U, V: [0, 1] \to \mathbb{R}^{d_z \times r}$ with $r < d_z$ be networks mapping the $\Delta$ to rank-$r$ matrices. We define
\begin{equation}
    \textcolor{red}{A_\Delta}=A_0 + U(\Delta) V(\Delta)^\top.
\label{eq:conditioned-transition}
\end{equation}
$V(\Delta)$ defines a preference-dependent $r$-dimensional subspace of the representation space, and $U(\Delta)$ specifies how this subspace contributes to the predicted future representation. 
Conceptually, the perturbation reads $r$ preference-relevant directions out of the current representation and writes $r$ corrections into the predicted future. Sharing $A_0$ across preferences grounds the learned dynamics to a common baseline, preventing the transition structure from drifting unpredictably as $\Delta$ varies. The parameterization also reduces the number of preference-dependent outputs from $d_z^2$ to $2rd_z$, which aids scalability to higher-dimensional representations and mitigates overfitting when the preference axis is sparsely sampled. Combined with $\textcolor{blue}{\psi_\Delta}$, this defines the conditioned anchor encoder $\phi_\Delta(s) = \textcolor{red}{A_\Delta}\textcolor{blue}{\psi_\Delta}(s)$, completing the conditioned logit $S_{ij}^{\Delta}$ in ~\eqref{eq:conditioned-logit}. In practice, $F_\xi$, $U$, and $V$ all operate on a shared Fourier feature embedding of $\Delta$~\citep{fourier}. Full architectural details appear in Appendix~\ref{appendix:architecture}.

\subsection{Preference-Conditioned Contrastive Objective}
\label{sec:objective}

For planning, the representation structure must encode the likelihood of observing future states from an anchor state, given its preference context. Therefore, anchor-future pairs are scored under the preference label of the anchor, such that for a batch of preference-labeled positive pairs $\{(s^i, s_+^i, \Delta_i)\}_{i=1}^B$, the conditioned logit is defined as $S_{ij}^{\Delta_i}$ using~\eqref{eq:conditioned-logit}. Positive pairs are sampled following~\citet{ivi}, with each pair inheriting $\Delta_i$ of its source trajectory. This ensures that positive pairs $(s^i, s_+^i)$ are pulled together under their shared preference, while negative pairs are pushed apart under the anchor preference. Extending the symmetrized InfoNCE loss with preference conditioning yields
\begin{equation}
    \mathcal{L}_{\text{\method}} = -
    \sum_{i=1}^B\left\{ \log\left(\frac{S_{ii}^{\Delta_i}}{\sum_{j =1,\, j \neq i}^B S_{ij}^{\Delta_i}}\right)+\log\left(\frac{S_{ii}^{\Delta_i}}{\sum_{j =1,\,j \neq i}^B S_{ji}^{\Delta_j}}\right) \right\}.
\label{eq:conditioned-info-nce}
\end{equation}
Minimizing $\mathcal{L}_{\text{\method}}$ jointly optimizes the encoder parameters, FiLM generator, low-rank modulation networks, and shared matrix $A_0$, which together define the planner $\mathcal{P}$ from Section~\ref{sec:problem-setup}. 
Under the same assumptions made by~\citet{ivi}, for any sampled pair $(s^i, s^j_+)$ and preference $\Delta \in [0, 1]$, the optimal conditioned similarity satisfies
\begin{equation}
    \big(S^{\Delta}_{ij}\big)^\star \propto \frac{p_\gamma(s_+^j \mid s^i, \Delta)}{p(s_+^j)},
\label{eq:conditioned-density-ratio}
\end{equation}
as shown in Appendix~\ref{appendix:density-ratio-assumption}. This is the conditioned analogue of~\eqref{eq:density-ratio}, preserving the density-ratio property of the base contrastive planner. The representation space now encodes how likely it is to observe a given future state from a starting state, under the user-preference set at runtime. This distinction from~\eqref{eq:density-ratio} enables behavioral modulation while keeping inference-driven planning tractable under \method. \citet{ivi} further assume that the marginal distribution over representations is isotropic Gaussian, which we argue remains consistent with our approach in Appendix~\ref{appendix:gaussianity-assumption}. At inference, a query $\mathbf{q} = (s_0, s_g, \Delta)$ is resolved by computing $\textcolor{red}{A_\Delta}$ and the conditioned endpoints, solving the Gauss-Markov posterior for $z_{1:n}$, and nearest-neighbor decoding as outlined in Algorithm~\ref{alg:momo}. In practice, standard goal-reaching policies trained over short horizons can then be used to track these conditioned waypoints for a complete trajectory.

\begin{algorithm}[t]
\caption{\method: Preference-Conditioned Contrastive Planning. Color denotes conditioning of the \textcolor{blue}{representation geometry} and \textcolor{red}{latent prediction matrix}; black follows \citet{ivi}}
\label{alg:momo}
\begin{algorithmic}[1]
    \STATE \textbf{\textsc{\underline{Training}}}
    \vspace{2pt}
    \STATE \textbf{Input: } offline trajectories $\mathcal{D} = \{(\tau_i, \Delta_i )\}_{i = 1}^N$
    \FOR{each minibatch}
        \STATE  Sample preference-labeled positive pairs $\{(s^i, s_+^i, \Delta_i)\}_{i = 1}^B$ from discounted state-occupancy 
        \STATE \textcolor{blue}{Encode anchors and candidate futures with $\psi_{\Delta_i}$}
        \STATE \textcolor{red}{Compute conditioned prediction matrix $A_{\Delta_i}$ via~\eqref{eq:conditioned-transition}}
        \STATE Form similarities $S_{ij}^{\Delta_i}$ and update parameters by minimizing~\eqref{eq:conditioned-info-nce} under the norm constraint
    \ENDFOR
    \STATE \hrulefill
    \STATE \textbf{\textsc{\underline{Inference}}}
    \vspace{2pt}
    \STATE \textbf{Input: } query $\mathbf{q} = (s_0, s_g, \Delta)$, waypoint count $n$
    \STATE \hspace{1em} \textcolor{blue}{Encode endpoints $z_0 = \psi_\Delta(s_0),\; z_g = \psi_\Delta(s_g)$}
    \STATE \hspace{1em} \textcolor{red}{Generate conditioned prediction matrix $A_\Delta$ via~\eqref{eq:conditioned-transition}}
    \STATE \hspace{1em} Solve $z_{1:n} = \Lambda_\Delta^{-1} \eta_\Delta$, where $\Lambda_\Delta, \eta_\Delta$ follow~\citep{ivi} with $\textcolor{blue}{\psi_\Delta},\textcolor{red}{A_\Delta}$ replacing $\psi,A$
    \STATE \hspace{1em} Decode $z_{1:n}$ to $s_{1:n}$ via nearest-neighbor retrieval under $\textcolor{blue}{\psi_\Delta}$ in $\mathcal{D}_{\mathrm{held}}$
    \RETURN Preference-conditioned plan $s_{1:n}$
\end{algorithmic}
\end{algorithm}
\section{Experiments}
\label{sec:experiments}


We evaluate \method across a range of environments to address the following questions:
\begin{enumerate}[itemsep=1.5pt, leftmargin=24pt, topsep=1.5pt]
    \item[\textbf{Q1.}] Does \method produce temporally consistent plans by preserving the representation structure required for inference-driven planning?
    \item[\textbf{Q2.}] Does \method produce plans that respond meaningfully to the user preference, or does conditioning collapse to a single behavior mode?
    \item[\textbf{Q3.}] Does \method maintain effective preference-conditioned planning across diverse task settings, including higher-dimensional and longer-horizon tasks?
\end{enumerate}


\begin{figure*}[t]
    \centering
    \includegraphics[width=0.9\textwidth]{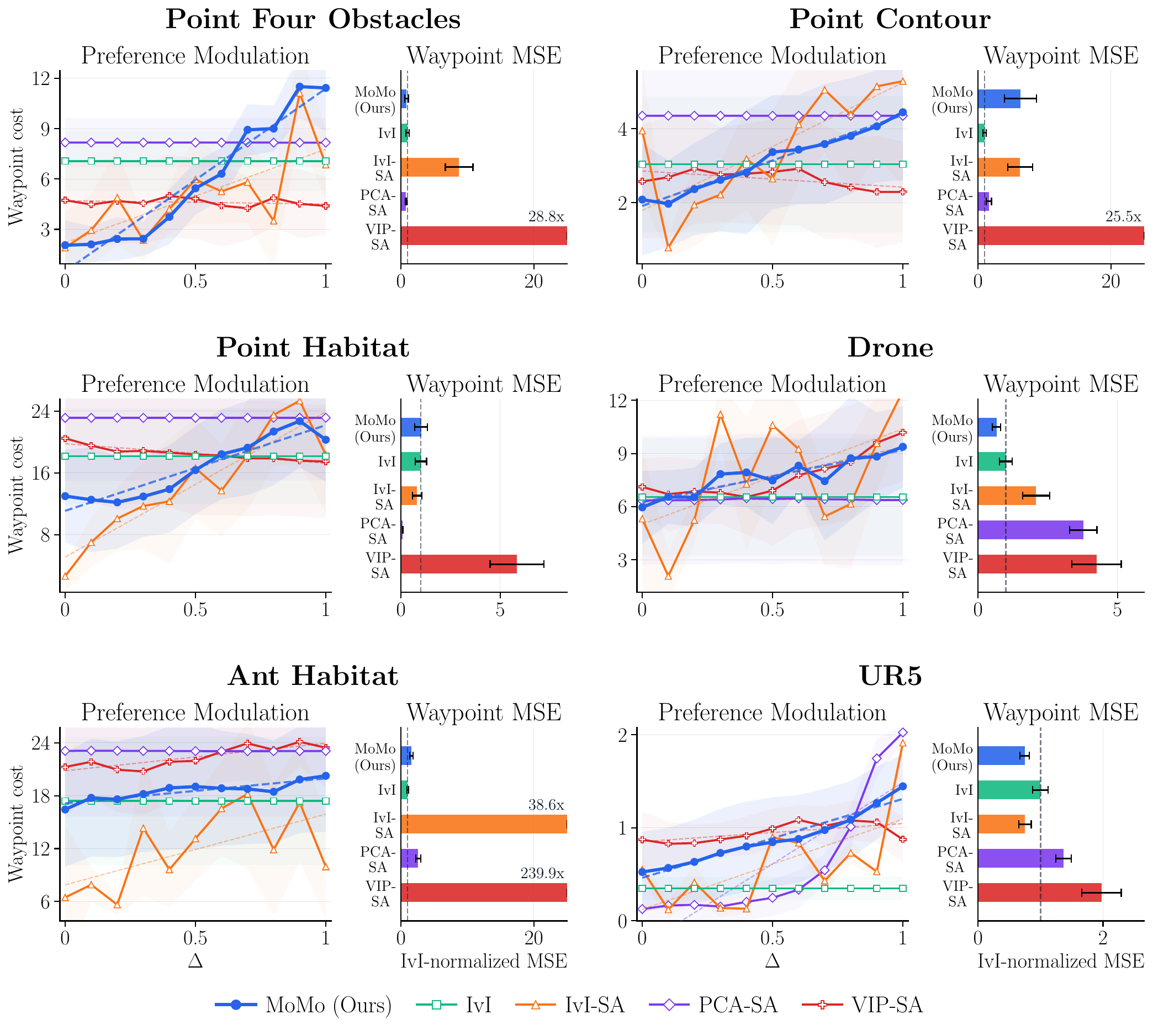}
    \vspace{-0.25cm}
    \caption{
        Preference modulation and planning fidelity across six environments. For each environment, the left panel plots waypoint cost as a function of preference $\Delta$; the right panel reports waypoint reconstruction MSE normalized by IvI's MSE (vertical reference line at $1.0$), where bars beyond $1.0$ indicate baselines that fit the dataset's waypoints worse than the preference-agnostic IvI. Dashed lines in the left panels show each method's modulation trend. Bars labeled with multipliers (e.g., $28.8\times$) extend beyond the displayed axis range. \method produces a near-monotonic trend between $\Delta$ and waypoint cost while maintaining MSE comparable to IvI; IvI is preference-agnostic and produces a flat trend; the SA baselines show inconsistent or non-monotonic responses with higher MSE.
    }
    \label{fig:quantitative_modulation}
\end{figure*}

\paragraph{Environments and Baselines.} We evaluate \method on six environments spanning four agents, summarized in Table~\ref{tab:datasets}, including three navigation tasks for a 2D \textit{Point} agent (an obstacle field, a Gaussian cost contour, and an indoor scene from AI Habitat~\citep{replica19arxiv, habitat19iccv, szot2021habitat, puig2023habitat3}), a quadcopter \textit{Drone} performing an exploration orbit around a central safe column, a quadruped \textit{Ant} navigating an AI Habitat scene, and a UR5 \textit{Manipulator} performing table-top pick-and-place with obstacle clearance. For each environment, we generate offline trajectory datasets at multiple discrete agent risk levels and annotate them with preference labels following Section~\ref{sec:preferences}. Dataset construction adapts D4RL~\citep{d4rl} and FSRL~\citep{fsrl}. Details about the datasets can be found in the Appendix~\ref{appendix:datasets}. We compare \method against four baselines. IvI~\citep{ivi} is the preference-agnostic baseline producing a single plan per start-goal query. The remaining baselines apply a state augmentation (SA) approach appending preference $\Delta_i$ as an additional observation dimension over three representation backbones: IvI-SA uses IvI's contrastive representations~\citep{ivi}, PCA-SA uses linear principal-component projections of raw observations, and VIP-SA uses representations from VIP~\citep{ma2023vipuniversalvisualreward} which encode temporal distances. Together, these isolate whether preference conditioning can be achieved via SA across various representation types.

\paragraph{Demonstrating Consistent Preference Modulation.} To evaluate \textbf{Q1} and \textbf{Q2}, we train \method across environments and query across a continuum of $\Delta$ values using Algorithm~\ref{alg:momo}. For \textbf{Q1}, we measure waypoint reconstruction MSE relative to IvI as an inference-driven planning reference. As shown in the right panel of Figure~\ref{fig:quantitative_modulation}, \method is able to achieve MSE comparable to IvI across most environments, while the SA baselines reach up to $28.8\times$ and $38.6\times$ higher on Point Four Obstacles and Ant Habitat, respectively. This indicates that joint conditioning preserves the representation structure required for planning by inference, whereas SA over alternative backbones disrupts it.

For \textbf{Q2}, we query start-goal pairs across multiple preferences and plot the sum of waypoint costs in the left panel of Figure~\ref{fig:quantitative_modulation}, observing a near-monotonic relationship between $\Delta$ and realized cost across all environments. \method's per-query variance is comparable to IvI's, indicating that modulation occurs at the individual query level rather than as a weak global average. Different SA backbones fail in qualitatively different ways. PCA-SA mostly collapses toward flat responses, indicating that linear projections lack sufficient structure to encode preference at all. IvI-SA produces noisy but correlated responses, showing that contrastive representations can encode preference under SA but yield unstable plans, which we examine further in Appendix~\ref{appendix:state_augmentation}. Finally, VIP-SA produces erratic responses with high MSE, suggesting that the temporal-distance representations are not preserved by SA-based conditioning. Figure~\ref{fig:teaser} illustrates the modulation of risk exposure for a given start-goal query, retrieving viable conditioned waypoint sequences which can be tracked by a standard goal-reaching policy to generate complete trajectories. Increasing $\Delta$ corresponds to more direct efficient routes, traded for greater realized risk. We further analyze how our conditioning method is able to achieve this modulation via deformation of the latent manifold in Appendix~\ref{appendix:additional_analysis}. \method is also able to demonstrate a mode-switching behavior in the left panel of Figure~\ref{fig:teaser} by recovering a different, safer approach for the same start and goal. 



\paragraph{Robustness across Tasks and Modalities.}
To evaluate \textbf{Q3}, we test \method along four robustness axes, with quantitative metrics for all six environments reported in Table~\ref{tab:risk_profile_metrics}. To test whether \method works across different environment risk structures, we evaluate the \textit{Point} agent in two longer-horizon settings, Point Contour with dense Gaussian cost-field contours and Point Habitat with an indoor scene rendered through AI Habitat. As shown in Figure~\ref{fig:quantitative_modulation}, \method maintains a Spearman correlation $\rho \geq 0.91$ between $\Delta$ and cost on both, with the smoothest modulation curves of any baseline. To test the inference-driven advantage of learning latent dynamics from a reduced feature set, we train the \textit{Drone} task on a 6-dimensional subset of the 18-dimensional observation space (translational positions and velocities only). Here, \method continues to modulate plans monotonically ($\rho = 0.87$), while PCA-SA's curve in Figure~\ref{fig:quantitative_modulation} collapses toward a single behavioral mode. This result highlights the performance of \method's joint conditioning approach in maintaining both temporal and preferential consistency on a reduced feature set where SA-based methods cannot.

To test representational capacity in high dimensions, we evaluate the 29-dimensional \textit{Ant} quadruped navigating an AI Habitat scene. Figure~\ref{fig:ant_headline} shows the resulting plans transitioning from cautious detours around obstacles at low $\Delta$ to direct routes at high $\Delta$. While VIP-SA achieves a marginally higher correlation ($\rho = 0.87$ vs. $0.82$), its waypoint MSE is significantly higher than IvI's, indicating that the plans themselves are not temporally consistent, whereas \method achieves the smoothest modulation curve across all baselines. While \method's correlation drops from $\rho = 0 .99$ on the 2-D Point environment to $\rho = 0.82$ here, the modulation remains monotonic and the waypoint MSE remains comparable to IvI. This graceful degradation across an order of magnitude in observation dimensionality suggests the approach is amenable to scaling. Finally, to test that \method does not depend on the specific choice of preference variable, we replace the CVaR-based preference on the UR5 \textit{Manipulator} with maximum end-effector height as a clearance proxy over a gate obstacle. As Figure~\ref{fig:ur5_headline} illustrates, \method recovers safe trajectories that maintain clearance above the divider, transitioning to aggressive plans passing directly through the gate as $\Delta$ increases, achieving the highest correlation as shown in Table~\ref{tab:risk_profile_metrics}. These analyses demonstrate that \method effectively performs preference-conditioned planning in complex settings, and is robust across a range of modalities.

\section{Conclusion}
\label{sec:conclusion}
We introduced \method, a preference-conditioned contrastive planning framework that jointly conditions the representation geometry and latent transition matrix, enabling inference-time control over user preferences while preserving the planning structure of CRLP. By conditioning at the logit level using a preference context statistic that summarizes behavioral tendencies within each trajectory, we preserve the density-ratio encoding in the latent structure required by CRLP. Across six environments and four agent embodiments, FiLM-based encoder modulation combined with low-rank modulation of the transition matrix produces temporally consistent plans with strong preference-cost correlation across diverse task settings including effective scaling to high-dimensional and longer-horizon tasks. More broadly, our work shows that representation-level conditioning, as opposed to policy- or cost-level conditioning, can support preference-conditioned planning at scale. Despite these encouraging results, there is further space for improvement upon the limitations outlined in Appendix~\ref{appendix:limitations}. One direction is to develop dual risk-and-goal-conditioned RL policies to track the conditioned waypoints, potentially mapping representations directly to actions to avoid noisy decoding. Another is to construct a general preference context embedding rather than a single scalar metric, to capture richer trajectory-level information. A third is to analyze the effects of network architecture on conditioned representation quality for planning.

\bibliographystyle{unsrtnat} 
\bibliography{refs}

\appendix
\newpage
\section{Conditioned Contrastive Theory}
\label{appendix:conditioning}

\subsection{Density Encoding in Learned Logits} 
\label{appendix:density-ratio-assumption}
The Gauss-Markov inference-driven planning method outlined in Section~\ref{sec:contrastive-prelim} requires learned logits to encode the probability ratio expressed in \eqref{eq:density-ratio}. This property must be preserved across the conditioning method. To prove that our training framework supports this, we begin by proving \eqref{eq:density-ratio} holds for the non-conditioned InfoNCE objective as done in~\citep{ivi}. We define the logits function $g(s,s_+)$ such that $g(s^i,s^j_+)=S_{ij}$. Start by considering the optimization problem which yields the optimal logits function,
\begin{equation}
\begin{aligned}
    \label{eq:opti_for_pres}
    \max_{g(s,s_+)}\lim_{B\rightarrow\infty}\mathop{\mathbb{E}}_{\{(s^i,s^i_+)\}_{i=1}^B\,\sim p(s,s_+)}\Bigg[\frac1B\sum_{i=1}^B&\Bigg\{ \log\left(\frac{g(s^i,s^i_+)}{\sum_{j=1,j\neq i}^B g(s^i,s^j_+)}\right)\\&+\log\left(\frac{g(s^i,s^i_+)}{\sum_{j=1,j\neq i}^B g(s^j,s^i_+)}\right) \Bigg\}\Bigg].
\end{aligned}
\end{equation}
Now consider the first term in the objective,
\begin{equation}
    \label{eq:preserve_def_J1}
    J_1(g)=\lim_{B\rightarrow\infty}\mathop{\mathbb{E}}_{\{(s^i,s^i_+)\}_{i=1}^B\,\sim p(s,s_+)}\Bigg[\frac1B\sum_{i=1}^B \log\Bigg(\frac{g(s^i,s^i_+)}{\sum_{j=1,j\neq i}^B g(s^i,s^j_+)}\Bigg)\Bigg].
\end{equation}
Taking the functional variation, the solution to the optimization problem can be extracted as follows.
\begin{align}
    \delta J_1(g)&=\lim_{B\rightarrow\infty}\mathop{\mathbb{E}}_{\{(s^i,s^i_+)\}_{i=1}^B\sim p(s,s_+)}\Bigg[\frac1B\sum_{i=1}^B\Bigg(\frac{\delta g(s^i,s^i_+)}{g(s^i,s^i_+)}-\frac{\delta\sum_{j=1,j\neq i}^Bg(s^i,s^j_+)}{\sum_{j=1,j\neq i}^Bg(s^i,s^j_+)}\Bigg)\Bigg] \nonumber \\
    &=\lim_{B\rightarrow\infty}\frac1B\sum_{i=1}^B\mathop{\mathbb{E}}_{s^i\sim p(s)}\left[\int\left(
    \frac{\delta g(s^i,s_+)}{g(s^i,s_+)}p(s_+\vert s^i)-k(s^i)\delta g(s^i,s_+)p(s_+)\right)ds_+
    \right] \nonumber \\
    &= \iint \delta g(s,s_+)\left[\frac{p(s_+|s)}{g(s,s_+)}-k(s)p(s_+)\right]p(s)dsds_+ 
    \label{eq:vanishing_integrand}
\end{align}
Note that in the second line, the expectation is rewritten over sampled anchors and integrated over futures, leaving a term
\begin{equation}
    k(s^i)=\lim_{B\rightarrow\infty}\mathop{\mathbb{E}}\left[\frac{B-1}{\sum_{j=1,j\neq i}^Bg(s^i,s^j_+)}\right]=\left[\int g(s^i,s_+)p(s_+)\,ds_+\right]^{-1},
    \label{eq:preserve_kfunc}
\end{equation}
which is independent of the future realizations in the infinite batch sampling limit. The first order condition of optimality 
requires the final integral to vanish in a pointwise manner for all arbitrary variations $\delta g(s,s_+)$. From the fundamental lemma of the calculus of variations~\citep{calculus_of_variations}, the bracketed term must therefore vanish for all anchor and future states. Absorbing $k(s)$ as a free normalization of logits~\citep{ivi}, we obtain the following optimal solution.
\begin{equation}
    g^*(s,s_+)\propto\frac{p(s_+\vert s)}{p(s_+)}
    \label{eq:preserve_res}
\end{equation}
Repeating this procedure for the second term in the original optimization problem $J_2(g)$ and noting that $p(s_+\vert s)p(s)=p(s\vert s_+)p(s_+)$ allows us to recover the same solution, ensuring consistency of the symmetrized objective. Using the function approximator $g(s,s_+)\approx\exp\{-\frac12\vert\vert A\psi(s)-\psi(s_+)\vert\vert^2\}$, this proves the statement in \eqref{eq:density-ratio}~\citep{ivi, fmicl, ma_and_collins}.
\par
Extending this to conditioned contrastive learning, define the conditioned logits function as $\hat g(s,s_+;\Delta)$. This leads to the following problem according to the form in \eqref{eq:opti_for_pres},
\begin{equation}
\begin{aligned}
    \label{eq:opti_for_pres_cond}
    \max_{\hat{g}(s,s_+;\Delta)}\lim_{B\rightarrow\infty}\mathop{\mathbb{E}}_{\{(s^i,s^i_+,\Delta_i)\}_{i=1}^B\,\sim p(s,s_+,\Delta)}\Bigg[\frac1B\sum_{i=1}^B&\Bigg\{ \log\left(\frac{\hat{g}(s^i,s^i_+;\Delta_i)}{\sum_{j=1,j\neq i}^B \hat{g}(s^i,s^j_+;\Delta_i)}\right)\\&+\log\left(\frac{\hat{g}(s^i,s^i_+;\Delta_i)}{\sum_{j=1,j\neq i}^B \hat{g}(s^j,s^i_+;\Delta_j)}\right) \Bigg\}\Bigg].
    \nonumber
\end{aligned}
\end{equation}
If we interpret anchor sampling as selecting an index at random from the concatenated tuple of all observed states in the dataset, we see that this implies both $s^i$ and $\Delta_i$. In other words, since the anchor state and preference context of each positive pair are sampled together, we can rewrite the anchor variable as $\hat{s}=[s^\top\;\Delta]^\top$ and accordingly an amended logit function $\doublehat{g}(\hat s,s_+)$. Note that this is not an augmentation of state since $s_+$ is unchanged, it is only for notational convenience. This yields the first component, built in the same manner as \eqref{eq:preserve_def_J1}, 
\begin{equation}
    \label{eq:preserve_def_J1_cond}
    J_1(\doublehat{g})=\lim_{B\rightarrow\infty}\mathop{\mathbb{E}}_{\{(\hat s^i,s^i_+)\}_{i=1}^B\,\sim p(\hat s,s_+)}\Bigg[\frac1B\sum_{i=1}^B \log\Bigg(\frac{\doublehat{g}(\hat s^i,s^i_+)}{\sum_{j=1,j\neq i}^B \doublehat{g}(\hat s^i,s^j_+)}\Bigg)\Bigg].
    \nonumber
\end{equation}
Note that sampling from $p(s,s_+,\Delta)$ as done in the original optimization problem is equivalent to sampling from $p(\hat s, s_+)$ following the reasoning outlined above. This translates to a functional variation, analogous to \eqref{eq:vanishing_integrand},
\begin{equation}
    \delta J_1(\doublehat{g}) = \iint \delta \doublehat{g}(\hat s,s_+)\left[\frac{p(s_+|\hat s)}{\doublehat{g}(\hat s,s_+)}-\doublehat{k}(\hat s)p(s_+)\right]p(\hat s)d\hat sds_+,
    \nonumber
\end{equation}
where $\doublehat{k}(\hat s)$ follows the definition corresponding to the formulation in \eqref{eq:preserve_kfunc}. Once again, from variational calculus~\citep{calculus_of_variations}, the bracketed term must vanish pointwise, yielding the proportionality of equivalent form to \eqref{eq:preserve_res},
\begin{equation}
    (\doublehat{g})^*(\hat s,s_+)\propto\frac{p(s_+\vert \hat s)}{p(s_+)}
    \label{eq:preserve_res_doublep}
    \nonumber
\end{equation}
Reverting back to the original conditioned logit definition, we have 
\begin{equation}
    (\hat{g})^*(s,s_+; \Delta)\propto\frac{p(s_+\vert s,\Delta)}{p(s_+)}
    \label{eq:preserve_res_cond},
\end{equation}
under the assumption that the sampling process is unchanged with the addition of a preference context variable. This is precisely the conditional probability ratio we intend to encode in the representation space, capturing the probability of observing a future state given both the anchor state and preference context. Repeating this procedure for $J_2(\doublehat{g})$ using the same vector augmentation of the anchor yields
\begin{equation}
    (\hat{g})^*(s,s_+; \Delta)\propto\frac{p(s,\Delta\vert s_+ )}{p(s, \Delta)},
    \label{eq:preserve_res_cond_j2}
\end{equation}
which is equivalent to the first proportionality result by Bayes' theorem. As such, we preserve the required latent dynamics for inference-driven planning, while reflecting a preference directly within the representation space geometry as intended.

\subsection{Gaussianity of Representations}
\label{appendix:gaussianity-assumption}
A second key assumption underlying CRLP is that the marginal distribution over representations is isotropic Gaussian~\citep{ivi}. In our formulation, we require this assumption to hold for all conditioned latent spaces. In other words, whilst~\citet{ivi} assumes $\psi\sim\mathcal{N}(\psi\,;\mu=0,\Sigma =c\cdot I)$, we take the point-wise version $\textcolor{blue}{\psi_\Delta}\sim\mathcal{N}(\textcolor{blue}{\psi_\Delta}\,;\mu=0,\Sigma =c\cdot I)$, $\forall\, \Delta$. Using the representation norm constraint enforces this in expectation, implying that the variance of the distribution could fluctuate over the preference continuum. We recognize that under conditioning, the density-ratio encoding and isotropic Gaussian assumptions face larger logit function approximation error as well as sampling error from the sparser joint state-preference space. However, we find empirically that planning performance remains consistent with~\citep{ivi}, generating representations that keep inference-driven planning tractable, as evidenced in Section~\ref{sec:experiments}.

\subsection{Planning Using Contrastive Representations} 
\label{appendix:contrastive-planning}
The distribution over intermediate latent waypoints $z_{1:n}$ from start and goal representations $z_0, z_g$, can be derived from the posterior 
\begin{equation}
    p(z_{1:n} \mid z_0, z_g) \propto p(z_g \mid z_n) \prod_{k=1}^n p(z_k \mid z_{k - 1}),
\label{eq:planning-posterior}
\end{equation}
in combination with~\eqref{eq:latent-dynamics}~\citep{ivi}. This yields the Gaussian distribution $z_{1:n}\sim\mathcal{N}(z_{1:n}\,;\mu= \Lambda^{-1} \eta, \Sigma=\Lambda^{-1})$, where   
\begin{gather}
    \Lambda = \text{tridiag}\left(-A,\,\frac{c}{c+1}A^\top A+\frac{c+1}{c}I,\,-A^\top\right), \\
    \eta =  \begin{bmatrix}
                z_0^\top A^\top & \mathbf{0} & \dots & \mathbf{0} & z_g^\top A
            \end{bmatrix}^\top.
\end{gather}
Conditioned waypoints are resolved by redefining all representations as $z_{(.)}=\textcolor{blue}{\psi_\Delta}(s_{(.)})$ and using $\textcolor{red}{A_\Delta}$ in place of $A$.

\section{Further Studies}
\begin{figure*}[t!]
    \centering
    \includegraphics[width=\textwidth]{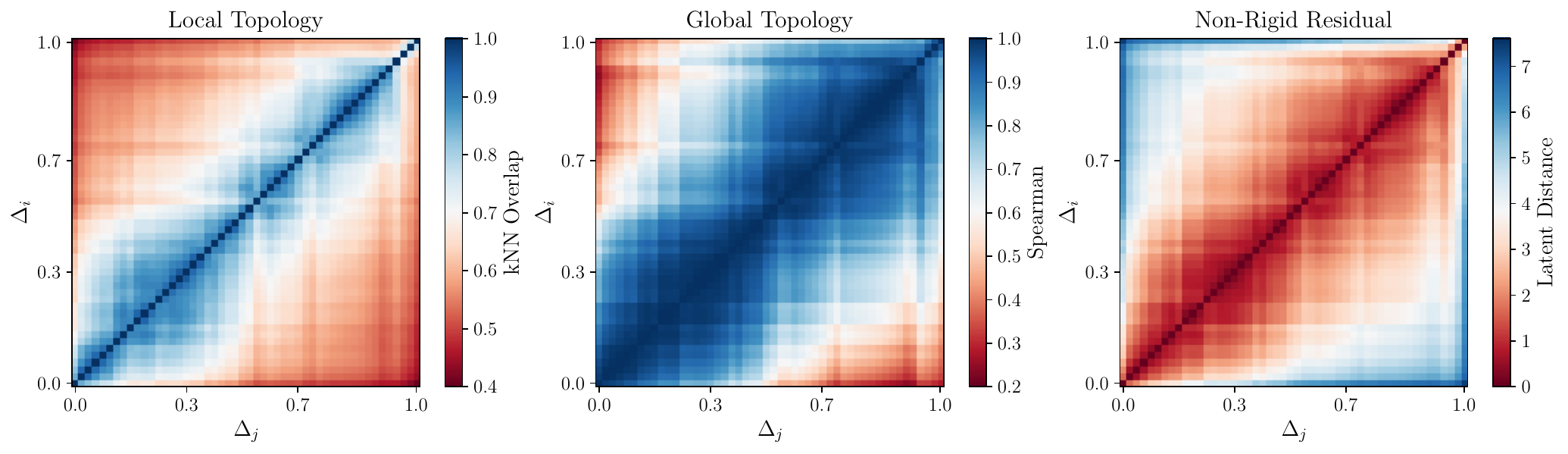}
    \vspace{-0.25cm}
    \caption{Matrix summaries of representation space manifold deformation for Point Four Obstacle environment under preference-conditioning. Plots characterize local changes via kNN overlap (left), global changes via Spearman coefficients (middle), and non-rigid transformation effects using residual latent distances (right).
    }
    \label{fig:manifold_deformation}
\end{figure*}
\subsection{Representation Space Deformation}
This section presents some further analysis of how conditioning affects the representation space structure. To characterize the deformation of the latent manifold, we embed a random collection of states across a range of preference contexts and consider different relation metrics between these conditioned representations. 

To understand how the local topology of the latent space changes, we apply $k$-nearest-neighbors (kNN) in the representation space to a subset of states for a range of $\Delta$ values. For every $(\Delta_i,\Delta_j)$ pair, we then compute the local overlap, defined as the average proportion of the $k$ neighbors which are shared between the two $\Delta$ values. Values closer to one indicate that local neighborhoods are highly similar, whilst a value of zero implies the local topology has changed completely. An example of the resultant matrix for the four obstacle environment is provided in the left panel of Figure~\ref{fig:manifold_deformation}, where we observe strong local overlap for similar preference values, which then smoothly decreases as the difference between $\Delta_i$ and $\Delta_j$ widens. Importantly, the lowest overlap level, occurring at the extremities of zero and one, is non-negligible at approximately 40\%. This captures temporal relations in regions of the representation space shared across all preference contexts.

For measuring global changes in the latent space, we compute the distance of randomly sampled state pairs for a range of $\Delta$ values and take the Spearman correlation between these distances, as shown in the middle panel of Figure~\ref{fig:manifold_deformation}. Values closer to one indicate that the global distance ranking is preserved. We again observe the dissipation of this similarity further from the diagonal entries and the strongest global deformation of the space for highly mismatched preferences.

To motivate the source of these deformations, we further embed a set of states at each preference value. For every $(\Delta_i, \Delta_j)$ pair, the two representation sets are centered, the best orthogonal alignment is then identified, and then the remaining RMS mismatch is reported in the right panel of Figure~\ref{fig:manifold_deformation}. A value of zero indicates that the deformation is a pure translation and rotation, so could be learned as a linear transformation. We observe that as the preferences differ, the residual rises and this rigid transformation is insufficient to describe the deformation. Instead, non-linear sources of preference-dependent warping are introduced. This further validates the choice of intermediate feature modulation in the conditioned encoder architecture. Together, these metrics point towards a healthy pattern in continuous preference modulation via deformation of the latent manifold.

\begin{figure*}[t!]
    \centering
    \includegraphics[width=0.65\textwidth]{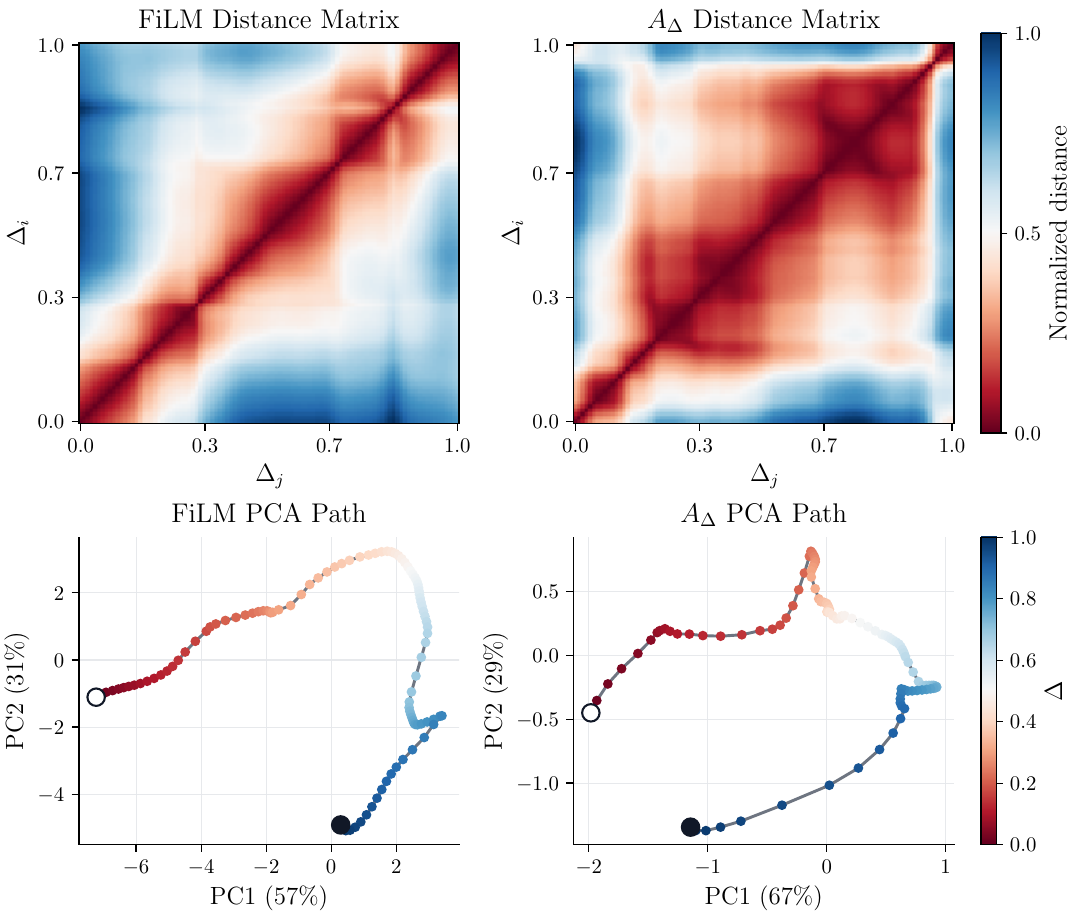}
    \vspace{-0.25cm}
    \caption{Summary of FiLM and transition matrix conditioning mechanism parameters for Ant Habitat environment. Across the preference spectrum, panels show the $\|.\|_2$ distance between FiLM generator outputs (top left), Frobenius norm $\|.\|_{F}$ of the conditioned transition matrix (top right), path of FiLM generator outputs across the first two principal components (bottom left), and the path of the flattened transition matrix elements across the first two principal components (bottom right).
    }
    \label{fig:film_Adelta}
\end{figure*}

We also analyze the FiLM and low rank modulated transition matrix which produce these effective deformations. Top panels of Figure~\ref{fig:film_Adelta} demonstrate how the FiLM generator's output parameters and perturbed matrix elements remain similar for close $\Delta$ values, and gradually separate with changing preferences. We visualize the path of these conditioning parameters in the PCA-projected space. This yields a continuous path consistent across the two conditioning mechanisms 
and points to the overall stability of the joint conditioning mechanism employed in \method.

\label{appendix:additional_analysis}

\subsection{State Augmentation Comparison}
\label{appendix:state_augmentation}

A candidate conditioning mechanism for preference modulation of planning behavior is state augmentation, which is used as a comparison baseline. In this approach, we define a new state $\tilde s=[s^\top \,\, \Delta]^\top$ such that $\tilde s_+=[s_+^\top \,\, \Delta_+]^\top$. Note that this is not equivalent to the rewriting of only the anchor as $\hat s=[s^\top \,\, \Delta]^\top$ which is done to perform a single marginalization step in the proof provided in Appendix \ref{appendix:density-ratio-assumption}. For the IvI-SA baseline, simply substituting the augmented state pair $(\tilde s, \tilde s_+)$, into the sequence of steps provided from~\eqref{eq:opti_for_pres} to~
\eqref{eq:preserve_res}, then unpacking the augmentation in the final step, we arrive at the density ratio
\begin{equation}
    \label{eq:density-ratio-augmentation}
    (\tilde g)^*(\tilde s, \tilde s_+)\propto \frac{p(s_+,\Delta_+\mid s,\Delta)}{p(s_+,\Delta_+)}.
    \nonumber
\end{equation}
This ratio captures the odds of seeing a future in its own distinct preference context given a starting state in another preference context. Note that $s_+$ represents any possible future following marginalization in~\eqref{eq:vanishing_integrand}, not strictly a positive one. Since positive pairs are sampled from the same trajectory, this would imply that $(\tilde g)^*$ is only non-zero when $\Delta=\Delta_+$. This risks collapsing the representation space since the appended preference context can now be used independently as a way of distinguishing positive and negative pairs, offering a degenerate solution to the optimization problem if the contrastive signal from remaining states is weak.

The learned Guass-Markov distribution therefore admits intermediate waypoints which share the same preference context as the augmented start and goal. The key distinction here is that a single unified representation space has been learned as opposed to a conditioned deformation. 

It is critical that with this formulation, state augmentation is able to encode precise temporal relations in the latent space. By inspection of Figure~\ref{fig:quantitative_modulation}, we observe large waypoint MSE using state augmentation. This is obtained alongside weaker or unstable trends between user preference and realized plan costs, indicated by lower Spearman correlations as well as considerably higher roughness in Table~\ref{tab:risk_profile_metrics}. If able to span a wider range of plan costs, it does not do so smoothly across the preference continuum. Therefore, across the metrics outlined in Table~\ref{tab:risk_profile_metrics}, \method demonstrates a consistent modulation of behaviors offering greater robustness across environments than the profiled baselines.

Under the injection of trajectory-level information into the augmented state, such that two identical original states can now lie on separate planes, state augmentation also becomes more susceptible to coverage issues in the held-out set for decoding, since every candidate state must be seen in the desired preference context.

\begin{table*}[t]
\centering
\small
\setlength{\tabcolsep}{4pt}
\renewcommand{\arraystretch}{1.15}
\caption{%
  Preference modulation profile metrics comparing \method to state augmentation baselines. $\Delta C$ measures endpoint contrast as the difference in total plan cost between $\Delta=1$ and $\Delta=0$ queries.
  $\rho$ is Spearman correlation between $\Delta$ and mean cost curve.
  $R_{\mathrm{norm}}$ is cost-curve roughness defined as the sum of second-order finite differences in cost normalized by $|\Delta C|$.
}
\label{tab:risk_profile_metrics}
\vspace{15pt}

\begin{tabular}{@{}l cccc cccc cccc@{}}
\toprule
\multirow{2}{*}{Environment}
  & \multicolumn{4}{c}{$\Delta C\ \uparrow$}
  & \multicolumn{4}{c}{$\rho\ \uparrow$}
  & \multicolumn{4}{c@{}}{$R_{\mathrm{norm}}\ \downarrow$} \\
\cmidrule(lr){2-5}\cmidrule(lr){6-9}\cmidrule(lr){10-13}
  & \textbf{Ours} & IvI$^*$ & PCA$^*$ & VIP$^*$ & \textbf{Ours} & IvI$^*$ & PCA$^*$ & VIP$^*$ & \textbf{Ours} & IvI$^*$ & PCA$^*$ & VIP$^*$ \\
\midrule
\shortstack[l]{Point Four\\Obstacles} & \textbf{9.37} & 4.94 & -0.01 & -0.34 & \textbf{0.99} & 0.77 & -0.95 & -0.32 & \textbf{1.32} & 7.75 & 2.70 & 13.35 \\
Point Contour & \textbf{2.36} & 1.34 & 0.00 & -0.28 & \textbf{0.99} & 0.85 & -0.81 & -0.57 & \textbf{0.78} & 10.14 & 2.78 & 5.56 \\
Point Habitat & 7.29 & \textbf{15.71} & 0.01 & -3.00 & 0.91 & \textbf{0.96} & 0.67 & -0.98 & \textbf{1.42} & 2.23 & 10.64 & 0.88 \\
Drone & 3.42 & \textbf{7.21} & 0.03 & 3.07 & \textbf{0.87} & 0.60 & 0.28 & 0.80 & 3.08 & 5.72 & 15.04 & \textbf{1.27} \\
Ant Habitat & \textbf{3.82} & 3.49 & -0.01 & 2.19 & 0.82 & 0.63 & -0.65 & \textbf{0.87} & \textbf{1.64} & 20.16 & 22.08 & 4.76 \\
UR5 & 0.92 & 1.36 & \textbf{1.90} & 0.00 & \textbf{1.00} & 0.57 & 0.97 & 0.72 & \textbf{0.29} & 4.63 & 0.67 & 225.90 \\
\addlinespace[2pt]
\bottomrule
\end{tabular}%
\par\smallskip
{\footnotesize\emph{Notes.}\enspace
\textit{Ours} = MoMo;\enspace $^*$ denotes the corresponding state-augmentation baseline.
Plain IvI is omitted because it is non-responsive ($\Delta C =0$). All environments use 12 waypoints apart from UR5 which uses 6.
}
\end{table*}

\section{Limitations}
\label{appendix:limitations}
In this section, we outline key limitations of our approach, both originating from preference conditioning and inherited from the inference-driven planning framework we build upon~\citep{ivi}.

Firstly, conditioning representations on $\Delta$ increases the computational complexity of a single training step from $\mathcal{O}(B)$ in~\eqref{eq:infonce} to $\mathcal{O}(B^2)$ in~\eqref{eq:conditioned-info-nce}. This arises from the need to embed states under each anchor's preference context, however this is not found to be computationally prohibitive. Furthermore, at inference time, nearest-neighbor retrieval mandates that the representations of observations in $\mathcal{D}_\text{held}$ be computed for the queried preference. Since we address preferences on a continuous scale, these embeddings cannot be pre-computed and cached. Whilst this reduces the runtime efficiency of the contrastive planner, a fast batched forward pass through the architecture is also not found to be prohibitive.

The safety of the decoded waypoint sequence is further reliant on a good coverage of the state space in the held-out set. It is also implicitly dependent on sufficient coverage of the joint preference and state space in the training set, since this is required to learn a generalizable conditioning mechanism. At inference time, if coverage in $\mathcal{D}_\text{held}$ is sparse, under the preference-dependent deformation of the latent space, nearest-neighbor retrieval could decode a safe latent waypoint back to an unsafe state. This limitation motivates an exploration of alternative decoding methods. 

We also inherit assumptions from \citet{ivi}, including that the learned logits encode the required density-ratio and that representations follow an isotropic Gaussian distribution. We justify these assumptions in Appendix~\ref{appendix:conditioning}; however, as in \citet{ivi}, it remains open how function approximation and sparse sampling errors introduced by conditioning affect \method's planning performance.

Finally, data-driven planning is exposed to the quality of existing datasets, which are difficult to acquire with dense coverage of the joint state-preference space. Furthermore, CRLP assumes all exploration of approaches to solving tasks has been performed, since no new temporal transitions external to the dataset can be extracted.

\section{Dataset Generation}
\label{appendix:datasets}

\begin{table}[t]
  \caption{Summary of training datasets with sufficient state-preference space coverage.}
  \label{tab:datasets}
  \centering
  \small
  \begin{tabular}{cccccc}
    \toprule
    \textbf{Environment} & \makecell{\textbf{Collection} \\ \textbf{framework}} & \makecell{\textbf{Risk} \\ \textbf{levels}} & \makecell{\textbf{Observation} \\ \textbf{dimension}} & \textbf{Trajectories} & \textbf{Transitions} \\
    \midrule
    Point Four Obstacle    & D4RL             & 3 & 4  & 16{,}557 & 3.31M \\
    Point Contour & D4RL             & 4 & 4  & 51{,}626 & 12.00M \\
    Point Habitat          & D4RL \& Habitat  & 3 & 4  & 29{,}488 & 7.50M \\
    Ant Habitat            & D4RL \& Habitat  & 3 & 29 & 20{,}186 & 7.50M \\
    Drone                        & FSRL             & 4 & 18 & 7{,}200  & 2.04M \\
    UR5                  & PyBullet         & 4 & 57 & 63{,}000 & 1.73M \\
    \bottomrule
  \end{tabular}
\end{table}

We construct trajectory datasets that provide coverage of the joint state-preference space required to learn preference-conditioned planning representations. For each of the six tasks, we collect rollouts at multiple discrete risk levels and annotate each trajectory with the CVaR-based preference statistic from Section~\ref{sec:preferences} except for the UR5 environment. 

For the \textit{Point} and \textit{Ant} agents, we adapt the D4RL collection framework~\citep{d4rl} to generate trajectories between randomized start and goal positions. The environment is discretized into a binary occupancy grid on which we perform Q-iteration to generate discrete waypoint sequences. Continuous trajectories between consecutive waypoints are then produced by a short-horizon goal-reaching policy. We use a stochastic PD-controller for the \textit{Point} agent, and a pre-trained SAC policy for the \textit{Ant} agent. Q-iteration uses discrete state $s$ corresponding to the position grid index, actions $a$ in the cardinal and intercardinal directions, deterministic transitions to the next state $s'$, goal $g$, and the reward


\begin{equation}
    R(s,a,g\,;w_\Delta) = -\left[ 1 + \max\bigl(0, \,d_b(s',g) -d_b(s, g)\bigr) + w_\Delta \left(1 - \frac{d_o(s)}{\theta}\right)\mathbbm{1}\!\bigl(d_o(s) < \theta\bigr)\right].
\label{eq:point-ant-reward}
\end{equation}

In the above, $d_b$ represents the shortest path distance between two cells (pre-computed via breadth-first search for all state pairs) and $d_o$ gives the distance to the nearest obstacle, for which a penalty is incurred within an influence radius $\theta$. By varying $w_\Delta$, the degree of risk-aversion baked into the Q-iteration waypoints is controlled. Rollouts are generated across discrete risk levels defined by distinct $w_\Delta$ values. We use a four obstacle map, open map with a dense Gaussian cost contour, and extract a map from Replica Dataset FRL Apartment 5 loaded from AI Habitat to model realistic environments. For annotation of the continuous trajectories, we use a risk cost function which increases linearly with obstacle proximity inside the influence radius.
\par
The dataset for the \textit{Drone} agent is produced using the data collection method and TRPO-Lagrangian policy for the \textit{OfflineDroneCircle-v0} environment from FSRL~\citep{fsrl}, trained at different target orbit radii. Preference annotation uses a risk cost which increases linearly with radial distance from a safe central column of radius $\theta$. 
\par
The \textit{Manipulator} environment consists of a tabletop pick-and-place task, requiring the robot to retrieve a cube from a start bowl and move it to the goal bowl. Between the bowls, there is a divider obstacle with a slot such that safer trajectories correspond to high clearance above the obstacle whilst more efficient, riskier motions slide through the divider. Trajectories are generated by simulation in PyBullet using a UR5 robot. To obtain the diverse set of preference behaviors, we apply varying obstacle inflation levels with Bi-directional Rapidly-exploring Random Trees~\citep{rrtconnect} for planning. Cost is computed from a margin-adjusted sigmoid function of vertical clearance between the end-effector and divider, weighted by the end-effector's $xy$-position, and wrapped to the unit interval.

\section{Model Training}
\label{appendix:architecture}

We build off of the unconditioned model presented in~\citep{ivi}, with key parameters summarized in Table \ref{tab:arch_params}. Since all conditioning modules, including the FiLM generator and two low rank modulation networks, would generate an intermediate representation of the preference context, we introduce a separate condition embedding network which feeds these three modules. This allows for an efficient shared condition embedding to inform the modulation mechanisms. 

Direct conditioning on the raw scalar $\Delta$ value may bottleneck performance due to the spectral bias of neural networks towards low frequency functions. This means that networks struggle to produce local fluctuations without affecting global behavior~\citep{spectral_bias_nn}, an important phenomenon for producing generalizable representations from sparse coverage of the joint state-preference space. This merits a richer representation of the conditioning variable, which we address using Fourier features~\citep{fourier}. Exact implementation details are contained in our codebase, to be released following reviews.

Whilst our derived results and those in~\citep{ivi} exclude the positive pair from the denominator of the InfoNCE objective, standard empirical implementations sum over all indices to yield a softmax distribution and bound loss. In line with prior work we also adopt this convention~\citep{ivi, betser2026infonceinducesgaussiandistribution}.

\begin{table}[t]
\centering
\setlength{\tabcolsep}{5.2pt}
\renewcommand{\arraystretch}{1.12}
\caption{%
  Model parameters used for \method.\ 
}
\label{tab:arch_params}
\begin{tabularx}{\linewidth}{@{}l>{\raggedright\arraybackslash}X@{}}
\toprule
\textbf{Component} & \textbf{Value} \\

\midrule
\multicolumn{2}{@{}l}{\textit{Encoder, predictor, and conditioning models}} \\[1pt]
\midrule
Representation dimension
  & \(d_z=4\) for \textit{Point} and \textit{Manipulator};
    \(d_z=6\) for \textit{Drone}; \(d_z=24\) for \textit{Ant} \\

Encoder network
  & Residual MLP with 5 residual blocks; width 64; Swish activations \\

FiLM generator network
  & MLP with 5 hidden layers; width 128; Swish activations \\

Condition embedding input features
  & \([\Delta,\sin(2\pi\Delta),\cos(2\pi\Delta),\sin(4\pi\Delta),\cos(4\pi\Delta)]\) \\

Condition embedding network
  & MLP with 2 hidden layers; width 16; Swish activations; 16-dimensional
    output linear embedding. \\

Low-rank \(A_\Delta\) network
  & Separate \(U_\Delta\) and \(V_\Delta\) MLPs each with 2 hidden layers;
    width 64; ReLU activations \\

Low-rank \(A_\Delta\) rank
  & \(r=2\) \\
CVaR level
  & \(\alpha=0.9\) \\

\midrule
\multicolumn{2}{@{}l}{\textit{Optimization}} \\[1pt]
\midrule
Batch size & 256 \\

Optimizer & Adam \\

Learning rate & \(3\times 10^{-4}\) \\

Global norm gradient clipping & 1.0 \\

Future sampling discount factor 
  & \(\gamma=0.9\) for \textit{Point, Ant, Manipulator}; $\gamma=0.8$ for \textit{Drone} \\

Maximum future horizon & 200 steps \\

Training step updates & 500,000 \\

Representation norm constraint target
  &  \(c=10\) \\

Dual descent initialization & 0.001 \\

\addlinespace[2pt]
\bottomrule
\end{tabularx}
\par\smallskip
\end{table}

We train our models on an Ubuntu 22.04 machine using a GeForce RTX 4080 GPU and Intel Core i9-14900K CPU, giving a training time of approximately 2 hours for \method.

\newpage

\end{document}